\documentclass[conference,letterpaper]{IEEEtran}
\usepackage[T1]{fontenc}
\usepackage{amsmath,amssymb}
\usepackage{booktabs}
\usepackage{url}
\usepackage{graphicx}
\makeatletter
\def\@fb@botlist{\@botlist}
\def\@fb@topbarrier{\suppressfloats[t]}
\def\FloatBarrier{\par\begingroup\let\@elt\relax
  \edef\@tempa{\@fb@botlist\@deferlist\@dbldeferlist}%
  \ifx\@tempa\@empty
  \else
    \ifx\@fltovf\relax
      \if@firstcolumn
        \clearpage
      \else
        \null\newpage\FloatBarrier
      \fi
    \else
      \newpage\let\@fltovf\relax
      \FloatBarrier
    \fi
  \fi
  \endgroup
  \@fb@topbarrier}
\makeatother
\graphicspath{{figures/}}
\title{A Probabilistic Lateral Motion Prediction Model\\with Improved Response Fidelity\\for Vehicle-in-the-Loop Simulation}
\IEEEoverridecommandlockouts
\author{%
  Xiangyang Ye\textsuperscript{1},
  Chaosheng Huang\textsuperscript{2},
  Jinwen Yang\textsuperscript{3},
  Zhengxian Chen\textsuperscript{2}, and
  Jun Li\textsuperscript{2}%
  \thanks{\textsuperscript{1}College of Automotive Engineering, Jilin University, Changchun, China.}%
  \thanks{\textsuperscript{2}School of Vehicle and Mobility, Tsinghua University, Beijing, China.}%
  \thanks{\textsuperscript{3}College of Mechanical Engineering, Tianjin University of Science and Technology, Tianjin, China.}%
}
\begin{document}
\maketitle
\begin{abstract}
Vehicle-in-the-loop (VIL) simulation requires a lateral dynamics model that accurately reproduces vehicle responses and reliably generates horizontal lateral-position and heading commands through recursive integration. This paper presents a probabilistic lateral motion prediction model with improved response fidelity. A two-degree-of-freedom vehicle model provides the physical prior, while a recurrent neural network learns correlated residual corrections to lateral acceleration and yaw rate. To maintain prediction fidelity during recursive motion generation, multistep rollout training feeds corrected responses into subsequent states and histories and directly penalizes accumulated lateral-velocity error. A conditional Gaussian characterizes response uncertainty, which is propagated to lateral-response, position, and heading predictions through augmented-state linearization. Evaluation results demonstrate improved lateral-response and integrated-motion prediction. Compared with the Deep Dynamics lateral adaptation, the proposed model reduces mean lateral-position root-mean-square error (RMSE) from 0.596 m to 0.483 m, an improvement of 19.0\%, while reducing mean lateral-velocity RMSE from 0.0469 m/s to 0.0441 m/s. These results demonstrate the effectiveness of combining physical priors, probabilistic residual learning, and multistep rollout training for response-faithful lateral motion prediction.
\end{abstract}

\section{Introduction}

\begin{figure}[!t]
\centering
\includegraphics[width=\columnwidth]{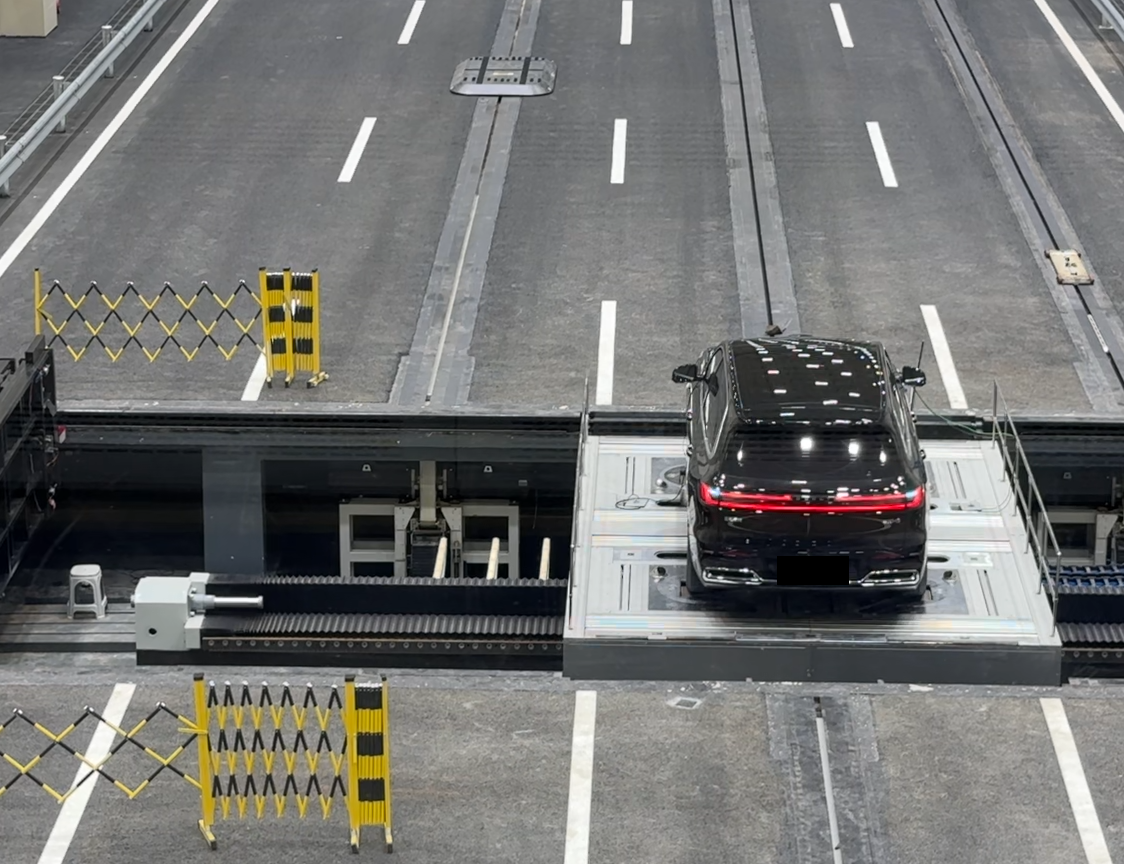}
\caption{Target vehicle-in-the-loop platform with horizontal lateral-position and heading command inputs.}
\label{fig:platform}
\end{figure}

Vehicle-in-the-loop (VIL) simulation connects a physical vehicle with a virtual driving environment, enabling the vehicle to interact with simulated driving scenarios~\cite{weiss2023highspeed}. In such systems, the vehicle dynamics model plays a central role in generating the motion commands supplied to the physical platform. The target platform considered in this work accepts horizontal lateral position $Y$ and heading $\phi$ as command inputs (Fig.~\ref{fig:platform}). Generating these commands requires a model that accurately reproduces vehicle lateral responses and maintains prediction fidelity during recursive motion integration. Therefore, the modeling objective extends beyond instantaneous response prediction to the reliable generation of integrated vehicle motion.

A fundamental challenge in lateral motion prediction arises from the coupling between lateral acceleration and yaw rate. In the vehicle-aligned coordinate system, lateral velocity evolves according to
\begin{equation*}
\dot v_y = a_y - ur,
\end{equation*}
where $a_y$ is lateral acceleration, $u$ is longitudinal speed, and $r$ is yaw rate. Persistent errors in the predicted acceleration and yaw-rate responses introduce an imbalance in $a_y-ur$, which accumulates into lateral-velocity error during recursive integration. This error subsequently affects the predicted vehicle state and the generated horizontal trajectory. Consequently, accurate single-step response prediction does not necessarily guarantee accurate free-running motion prediction, particularly when the model generates future states without measurement-based corrections. Maintaining response fidelity throughout recursive prediction is therefore essential for reliable lateral motion generation.

Physical vehicle models provide a structured description of the relationship between steering inputs, vehicle speed, and lateral motion. However, simplified models with fixed parameters cannot fully capture the response variations observed in practical driving conditions. Data-driven residual learning offers a complementary approach by learning discrepancies between physical predictions and measured vehicle responses. Existing hybrid models combine physical dynamics with learned corrections~\cite{hewing2018cautious,li2025residual,chen2024hybrid,miao2025residual}, while other approaches learn or adapt physical model parameters from observed data~\cite{deepdynamics,fang2025finetuning,ding2024drifting}. Nevertheless, improving local response accuracy alone does not necessarily ensure accurate integrated motion, particularly when prediction errors are repeatedly fed back into the dynamics model.

Reliable recursive motion prediction requires the model to account for the effects of its own predictions on subsequent states and response histories. Multistep rollout learning provides a means of optimizing prediction accuracy over recursively generated trajectories rather than isolated time steps~\cite{mohajerin2019multistep,tumu2023physics}. For lateral dynamics, directly penalizing accumulated lateral-velocity error provides a mechanism for constraining the imbalance between lateral acceleration and yaw rate during recursive prediction. In addition, vehicle responses exhibit variability that cannot be fully characterized by deterministic predictions. Probabilistic dynamics models provide a framework for representing such uncertainty and propagating it into future trajectories~\cite{chua2018pets}. Integrating probabilistic response modeling with recursive state propagation can therefore provide both predicted lateral motion and an associated measure of uncertainty.

In this paper, we develop a physics-guided probabilistic lateral motion prediction model for vehicle-in-the-loop simulation. A two-degree-of-freedom bicycle model provides the nominal physical response, while a recurrent neural network learns correlated probabilistic corrections to lateral acceleration and yaw rate. The model integrates the lateral-velocity dynamics using the corrected responses and sets the next yaw rate to its corrected value. It then integrates the predicted motion to obtain horizontal lateral position and heading. To improve prediction fidelity during extended free-running simulation, we introduce multistep rollout training with direct supervision of accumulated lateral-velocity error. Furthermore, the conditional response covariance is propagated through an augmented-state linearization to quantify uncertainty in both lateral responses and integrated vehicle motion.

The main contributions of this work are summarized as follows:
\begin{enumerate}
\item We develop a physics-guided probabilistic lateral motion prediction model that combines a two-degree-of-freedom vehicle model with recurrent residual learning. The model predicts correlated corrections to lateral acceleration and yaw rate and integrates the corrected responses to generate horizontal lateral position and heading.

\item We formulate a multistep rollout learning strategy that recursively feeds corrected responses and self-generated states back into the model. By directly penalizing accumulated lateral-velocity error, the proposed strategy improves response fidelity during free-running motion prediction.

\item We introduce an augmented-state covariance propagation method that maps probabilistic response uncertainty to lateral-response and integrated-motion predictions. The resulting framework provides both conditional-mean motion trajectories and associated prediction intervals for vehicle-in-the-loop simulation.
\end{enumerate}

\section{Related Work}
Hybrid vehicle models combine physical structure with learned corrections. Gaussian-process residuals have been used in autonomous racing~\cite{hewing2018cautious}, and low-dimensional residual models have been developed for driving control~\cite{li2025residual}. For lateral dynamics, Chen et al. combine a physical prior with an attention-based recurrent residual model~\cite{chen2024hybrid}; Miao et al. use a transformer to improve vehicle-dynamics fidelity~\cite{miao2025residual}. These approaches retain known motion relationships while learning discrepancies from measured responses.

An alternative is to learn the parameters of the physical model. Deep Dynamics estimates bounded physical coefficients from state and input histories and uses them within the vehicle equations~\cite{deepdynamics}. Physics-informed fine-tuning has also been studied for vehicle dynamics estimation~\cite{fang2025finetuning}, while Ding et al. investigate online neural vehicle modeling with unknown tire properties~\cite{ding2024drifting}. The present model instead learns additive, correlated corrections to lateral acceleration and yaw rate.

Recurrent multistep identification addresses prediction under feedback and history-based initialization~\cite{mohajerin2019multistep}. Tumu et al. propagate trajectory losses through vehicle dynamics for motion prediction~\cite{tumu2023physics}. Our formulation targets the acceleration imbalance $a_y-u r$ and its repeated integration into vehicle-aligned lateral velocity.

Probabilistic dynamics models, including PETS~\cite{chua2018pets}, describe response uncertainty and propagate it into trajectories. Proper scoring rules support distribution learning, while coverage and interval width characterize predictive performance~\cite{gneiting2007proper,gneiting2007calibration}. We combine a correlated response distribution with multistep lateral-dynamics learning and evaluate the resulting uncertainty in both responses and integrated motion.

\section{Method}
\subsection{Horizontal Motion and Physical Prior}
All motion considered here lies in a plane perpendicular to gravity. Fixed horizontal axes define $Y$ and $\phi$; horizontal axes rotating with vehicle heading define longitudinal speed $u$ and lateral velocity $v_y$. With the planar approximation $r=\dot\phi$,
\begin{align}
\dot v_y&=a_y-u r, & \dot\phi&=r,\\
\dot Y&=u\sin\phi+v_y\cos\phi.
\label{eq:kinematics}
\end{align}
Steering and longitudinal speed are prescribed inputs.

The physical prior is a linear two-degree-of-freedom bicycle model. Under the small-angle approximation, the front and rear lateral axle forces are
\begin{align}
F_{yf}&=C_f\left(\delta-\frac{v_y+l_f r}{u_c}\right),\\
F_{yr}&=-C_r\frac{v_y-l_r r}{u_c},
\label{eq:axle-forces}
\end{align}
where $\delta$ is the equivalent front-wheel steering angle in radians, $C_f,C_r$ are positive axle cornering stiffnesses, and $l_f,l_r$ are the distances from the center of mass to the front and rear axles. Lateral force and yaw moment balance give
\begin{align}
f_v=\dot v_y&=\frac{F_{yf}+F_{yr}}{m}-u r,\\
f_r=\dot r&=\frac{l_fF_{yf}-l_rF_{yr}}{I_z},
\label{eq:bicycle-dynamics}
\end{align}
where $m$ denotes vehicle mass, $I_z$ yaw moment of inertia, $l_f$ and $l_r$ the distances from the center of mass to the front and rear axles, and $C_f$ and $C_r$ the corresponding positive axle cornering stiffnesses. The quantity $u_c$ is the longitudinal speed used in the tire-slip expressions, regularized at low speed to avoid division by values close to zero.

Let $s_t=(v_{y,t},r_t)$ and $e_t=(\Delta t_t,\delta_t,u_t)$. Evaluating~\eqref{eq:axle-forces}--\eqref{eq:bicycle-dynamics} at $(s_t,e_t)$ gives the next-step physical response
\begin{align}
r^{\mathrm{phy}}_{t+1}&=r_t+\Delta t_t f_r(s_t,e_t),\\
a^{\mathrm{phy}}_{y,t+1}&=f_v(s_t,e_t)+u_t r^{\mathrm{phy}}_{t+1}.
\label{eq:physical-response}
\end{align}
Integrating the physical lateral-acceleration response gives $v^{\mathrm{phy}}_{y,t+1}=v_{y,t}+\Delta t_t f_v$. The learned model retains this response-to-state relationship.

\subsection{Probabilistic Response Correction}
\begin{figure*}[!t]
\centering
\includegraphics[width=\textwidth]{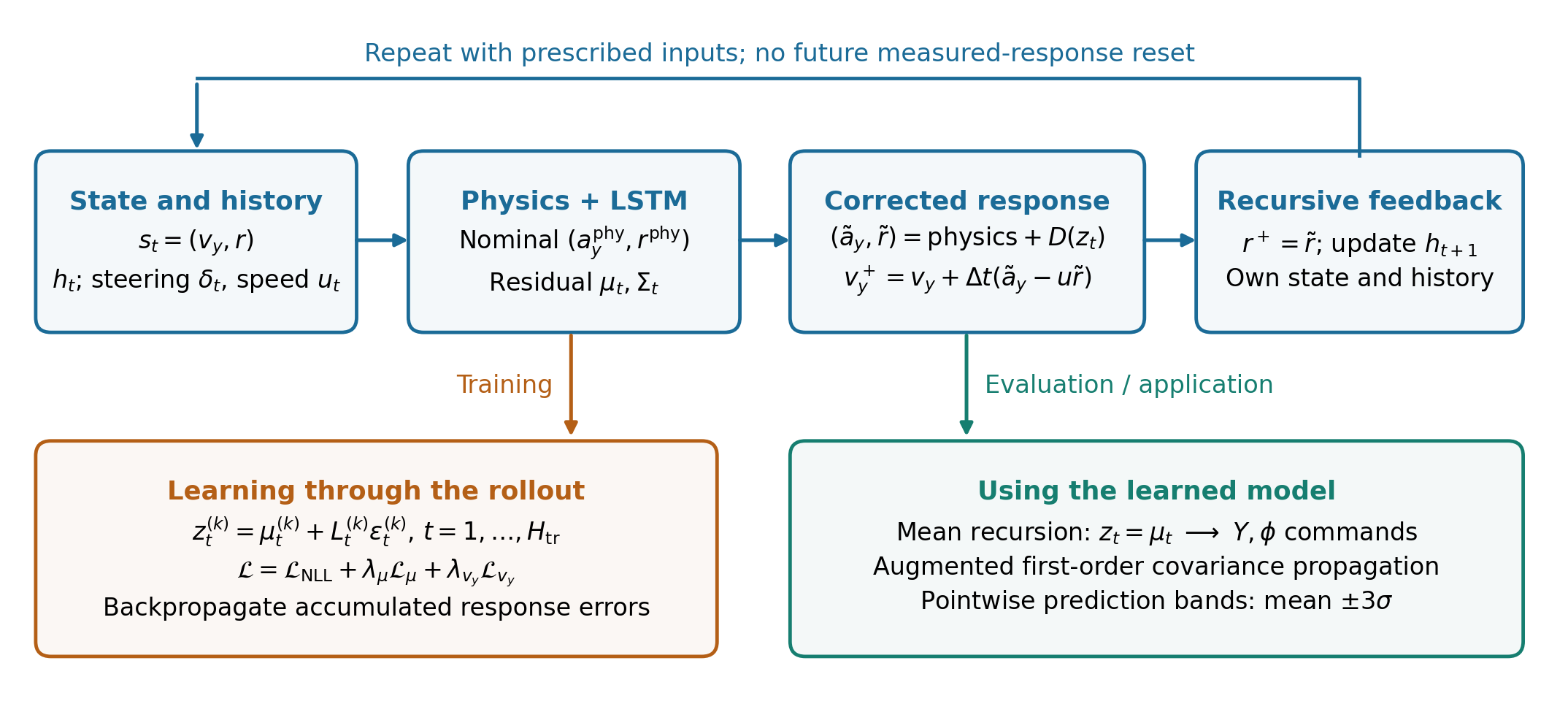}
\caption{Proposed lateral dynamics model. Corrected responses advance the state and history throughout training. Reparameterized rollout samples support learning; conditional-mean recursion and linearized covariance propagation yield horizontal motion predictions and prediction bands.}
\label{fig:method}
\end{figure*}

A 128-unit long short-term memory (LSTM) network encodes ten samples of steering, longitudinal speed, lateral acceleration, and yaw rate. The current inputs and physical response provide the prediction context. In normalized residual coordinates, the network outputs
\begin{equation}
z_t\mid h_t,c_t\sim\mathcal N(\mu_\theta,\Sigma_\theta),\qquad
\Sigma_\theta=\begin{bmatrix}\sigma_a^2&\rho\sigma_a\sigma_r\\\rho\sigma_a\sigma_r&\sigma_r^2\end{bmatrix}.
\label{eq:distribution}
\end{equation}
Positive standard deviations and bounded correlation parameterize the covariance. Let $D(z)=T(b+Sz)$ map normalized residuals to physical responses, using the training mean $b$ and diagonal scale $S$. Here $T$ is a diagonal scaling matrix that maps the normalized residual channels to the physical response units. The corrected response is
\begin{equation}
\begin{bmatrix}\tilde a_{y,t}\\\tilde r_t\end{bmatrix}
=\begin{bmatrix}a^{\mathrm{phy}}_{y,t+1}\\r^{\mathrm{phy}}_{t+1}\end{bmatrix}+D(z_t).
\end{equation}
The state advances according to
\begin{align}
r_{t+1}&=\tilde r_t,\\
v_{y,t+1}&=v_{y,t}+\Delta t_t(\tilde a_{y,t}-u_t\tilde r_t).
\label{eq:corrected-state}
\end{align}
The corrected responses also enter the next history window. Thus $\Delta a_y$ and $\Delta r$ are residuals relative to the physical responses; $\Delta r$ denotes a yaw-rate correction.

\subsection{Multistep Rollout Learning}
For training horizon $H_{\mathrm{tr}}$, each particle samples a reparameterized residual,
\begin{equation}
z_t^{(k)}=\mu_\theta^{(k)}+L_\theta^{(k)}\epsilon_t^{(k)},\quad
\epsilon_t^{(k)}\sim\mathcal N(0,I),\quad LL^\top=\Sigma.
\end{equation}
Each particle maintains its own state and response history throughout the rollout. The objective is
\begin{equation}
\mathcal L_{H_{\mathrm{tr}}}=\frac{1}{H_{\mathrm{tr}}K}
\sum_{t=1}^{H_{\mathrm{tr}}}\sum_{k=1}^{K}
\left(\ell_{\mathrm{NLL},t}^{(k)}+\lambda_\mu\ell_{\mu,t}^{(k)}+\lambda_{v_y}\ell_{v_y,t}^{(k)}\right).
\label{eq:training}
\end{equation}
The normalized target residual is computed relative to the physical response at each predicted state. The three terms are conditional negative log likelihood, weighted residual-mean MSE, and normalized lateral-velocity MSE. For normalized acceleration and yaw-rate residual errors $e_a$ and $e_r$, the mean term is $\ell_\mu=(0.2e_a^2+e_r^2)/2$. The state term is $\ell_{v_y}=((v_y^{(k)}-v_y^{\mathrm{ref}})/s_v)^2$, where $s_v$ is the training lateral-velocity standard deviation. We use four training particles, $\lambda_\mu=0.05$, and $\lambda_{v_y}=0.2$. Thus $\lambda_\mu$ weights response-mean accuracy, while $\lambda_{v_y}$ weights the integrated lateral-velocity error. All normalization statistics are estimated from training data.

An early error in $a_y-u r$ changes subsequent lateral velocity, the physical prediction, and the response history. Backpropagation through the rollout penalizes this accumulated effect. The $v_y$ term therefore connects local response correction with the recursive behavior required for motion generation.

Checkpoint selection uses a validation horizon $H_{\mathrm{val}}$:
\begin{equation}
\theta^\star(H_{\mathrm{val}})=\underset{e}{\operatorname{argmin}}\;
V_{H_{\mathrm{val}}}(\theta_e).
\label{eq:selection}
\end{equation}
Here $V$ evaluates the response objective, including NLL, along the conditional-mean recursion. The training horizon determines the gradients, whereas the validation horizon determines which trained parameters are retained.

\subsection{Horizontal Motion Prediction and Covariance Propagation}
The conditional-mean trajectory uses $z_t=\mu_\theta(h_t,c_t)$ at each step. Heading and lateral position follow by trapezoidal integration of yaw rate and horizontal velocity in~\eqref{eq:kinematics}. For trajectory visualization, longitudinal position follows $\dot X=u\cos\phi-v_y\sin\phi$ using the same integration rule. For the intended platform application, the resulting $Y/\phi$ predictions can be transmitted by UDP as tracking commands.

To evaluate predictive spread, we linearize the complete feedback transition about this conditional-mean trajectory. The 44-dimensional augmented vector $q_t$ contains $v_y$, $r$, $\phi$, $Y$, and the ten-sample input-response history. The transition $F$ includes the learned residual mean; its third argument is a zero-mean physical response disturbance. At the current mean state and history, this disturbance and its covariance are
\begin{equation}
\begin{aligned}
\eta_t&=TS(z_t-\mu_{\theta,t}),\\
Q_t&=\operatorname{Cov}(\eta_t)
     =TS\Sigma_{\theta,t}S^\top T^\top.
\end{aligned}
\label{eq:physical-covariance}
\end{equation}
Let $A_t$ and $G_t$ be the Jacobians of $F$ with respect to the augmented state and this disturbance, evaluated at $(\bar q_t,e_t,0)$. Then
\begin{equation}
\begin{aligned}
\bar q_{t+1}&=F(\bar q_t,e_t,0),\\
P_{t+1}&=A_tP_tA_t^\top+G_tQ_tG_t^\top .
\end{aligned}
\label{eq:covariance}
\end{equation}
The shared measured initialization is deterministic, so $P_0=0$. Output Jacobians map this covariance to the lateral response and pose quantities. Yaw acceleration is the backward difference of adjacent yaw rates, with their cross-time covariance retained in its propagated variance. The pointwise band for each quantity is $\bar y_t\pm3\sigma_t$, where $\sigma_t$ is its propagated marginal standard deviation.

\section{Offline Evaluation and Results}
\begin{table*}[!t]
\centering
\setlength{\tabcolsep}{3pt}
\caption{RMSE on six test maneuver windows. Each learned entry is the mean $\pm$ sample standard deviation over three training repetitions, after equal averaging of record-level RMSE. Physics is deterministic. $H_{\rm val}$ denotes the checkpoint-selection horizon, with each model retaining its own validation objective. Bold marks the lowest unrounded mean in each column.}
\label{tab:mean-accuracy}
\begin{tabular}{lccccc}
\toprule
Model & $H_{\rm val}$ & $v_y$ (m/s) & $r$ (deg/s) & $Y$ (m) & $\phi$ (rad) \\
\midrule
Physics & -- & $0.3094$ & $2.685$ & $1.238$ & $0.0508$ \\
Rollout $H_{\rm tr}=1$ & 100 & $0.2496\pm 0.0013$ & $4.400\pm 0.148$ & $0.934\pm 0.147$ & $0.0433\pm 0.0009$ \\
Rollout $H_{\rm tr}=5$ & 100 & $0.1290\pm 0.0100$ & $2.386\pm 0.179$ & $0.619\pm 0.046$ & $0.0231\pm 0.0018$ \\
Rollout $H_{\rm tr}=10$ & 100 & $0.1283\pm 0.0103$ & $1.647\pm 0.084$ & $0.602\pm 0.184$ & $0.0189\pm 0.0028$ \\
Rollout $H_{\rm tr}=20$ & 100 & $0.0833\pm 0.0108$ & $1.071\pm 0.049$ & $0.543\pm 0.087$ & $0.0167\pm 0.0016$ \\
Rollout $H_{\rm tr}=50$ & 100 & $0.0448\pm 0.0021$ & $0.915\pm 0.032$ & $0.543\pm 0.043$ & $0.0176\pm 0.0010$ \\
Rollout $H_{\rm tr}=100$ & 100 & $0.0441\pm 0.0042$ & $0.858\pm 0.012$ & $0.483\pm 0.036$ & $0.0161\pm 0.0010$ \\
Rollout $H_{\rm tr}=20$ + $L_{\rm pose}$ & 100 & $0.0810\pm 0.0087$ & $1.028\pm 0.039$ & $0.518\pm 0.050$ & $0.0166\pm 0.0011$ \\
Rollout $H_{\rm tr}=50$ + $L_{\rm pose}$ & 100 & $0.0481\pm 0.0049$ & $0.910\pm 0.039$ & $0.501\pm 0.028$ & $0.0163\pm 0.0001$ \\
Rollout $H_{\rm tr}=100$ + $L_{\rm pose}$ & 100 & $\mathbf{0.0437\pm 0.0016}$ & $\mathbf{0.856\pm 0.011}$ & $\mathbf{0.452\pm 0.033}$ & $\mathbf{0.0159\pm 0.0008}$ \\
Rollout $H_{\rm tr}=100$ w/o $L_{v_y}$ & 100 & $0.1347\pm 0.0047$ & $1.331\pm 0.106$ & $0.595\pm 0.061$ & $0.0196\pm 0.0014$ \\
Rollout $H_{\rm tr}=100$ + $L_{\rm pose}$ w/o $L_{v_y}$ & 100 & $0.1176\pm 0.0046$ & $1.286\pm 0.086$ & $0.555\pm 0.023$ & $0.0169\pm 0.0012$ \\
Chen24 & 1 & $0.2356\pm 0.0143$ & $4.413\pm 0.441$ & $1.047\pm 0.195$ & $0.0604\pm 0.0094$ \\
Chen24 & 100 & $0.2586\pm 0.0205$ & $3.648\pm 0.324$ & $0.797\pm 0.068$ & $0.0459\pm 0.0053$ \\
DeepDynamics lateral & 1 & $0.0476\pm 0.0057$ & $1.384\pm 0.147$ & $1.071\pm 0.307$ & $0.0331\pm 0.0048$ \\
DeepDynamics lateral & 100 & $0.0469\pm 0.0025$ & $1.064\pm 0.032$ & $0.596\pm 0.094$ & $0.0186\pm 0.0017$ \\
\bottomrule
\end{tabular}
\end{table*}

\subsection{Data and Evaluation}
Driving data were collected with an RT3000 system. The 46 records comprise 44 lane changes and two slaloms, partitioned by file into 33 training, seven validation, and six test records. All windows from a file stay in its partition; normalization uses training data only. Each learned configuration has three fresh training repetitions with 1,800 updates, batch size 1,024, and learning rate $10^{-3}$. Validation alone selects checkpoints before test evaluation. Heading and position references integrate recorded horizontal responses in the original local frame.

At the 0.01 s sampling interval, we compare observation-updated prediction with free simulation. The former predicts one step from the preceding measured state and ten measured history samples. Free simulations receive these measurements only at initialization, with common initial $v_y/r$ and reconstructed $Y/\phi$. Six test dynamic windows, fixed from recorded steering and yaw responses, span 4.13--4.83 s including recovery. A complete 24.03 s record additionally covers acceleration, a lane change near 100 km/h, and recovery; it is initialized once at 0.1 s after the ten-sample history. No subsequent measurement resets either free-simulation mode, and the initial point is excluded from scoring.

Dynamic and one-step RMSEs are averaged equally over the six records. Tables report mean $\pm$ sample standard deviation across three training repetitions. Complete-scene metrics instead concern the one fixed record. Figures use a predesignated repetition, without selecting the best run.

\begin{figure*}[!t]
\centering
\includegraphics[width=\textwidth]{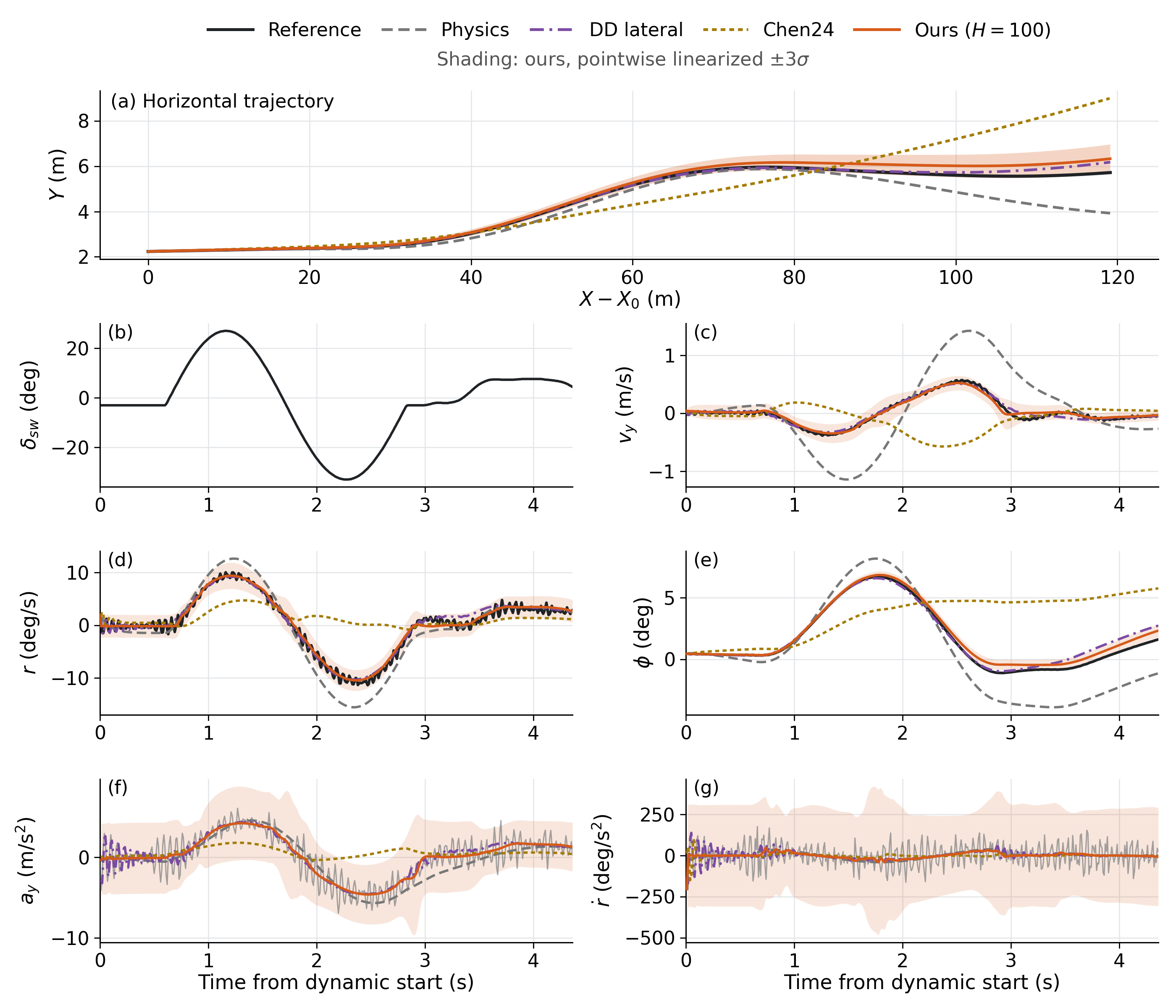}
\caption{Test lane change near 100 km/h, comparing physics, Chen24, DeepDynamics, and the rollout model. Top: horizontal $X$--$Y$ trajectories, with shading showing the rollout model's $Y$-direction distribution as mean $\pm3\sigma$ along the mean trajectory. Below: steering and lateral responses with aligned time axes; response shading shows the corresponding linearized mean $\pm3\sigma$. All models share the measured initialization; learned models use 100-step validation selection. The reference path integrates recorded horizontal motion; $X-X_0$ is displacement from the common dynamic start.}
\label{fig:scenario}
\end{figure*}

\subsection{Models and Checkpoint Selection}
The physical baseline shares the recorded inputs and measured initialization. Rollout variants use $H_{\mathrm{tr}}=1,5,10,20,50,100$, with pose supervision at 20, 50, and 100 steps and removal of $L_{v_y}$ at 100 steps, with and without pose supervision. Pose supervision adds mean-path $Y/\phi$ errors with weights 0.1 each and scales 1 m and 0.1 rad. All these ablations retain the same 100-step validation objective, including $\lambda_{v_y}=0.2$ and excluding pose terms.

The \emph{Chen24} baseline (Chen et al., 2024) combines temporal convolution, channel attention, a history-initialized 128-unit gated recurrent unit (GRU), and temporal attention to predict acceleration/yaw-rate residuals. It uses one-step weighted normalized response MSE plus the lateral-velocity term. The modules follow the motivation of~\cite{chen2024hybrid}; dimensions, fusion, and the residual interface are local choices, rather than an exact reproduction. The \emph{DeepDynamics lateral adaptation} uses a GRU, Mish activations, and Physics Guard parameter bounds~\cite{deepdynamics}, adapted to this vehicle with prescribed longitudinal speed. Training minimizes one-step unnormalized SI-state MSE. Baselines retain their native state updates; pose integration is trapezoidal for all models.

Paired selectors use $H_{\mathrm{val}}=1$ or 100 on the same saved training trajectories for both baselines and our $H_{\mathrm{tr}}=1$, 100, and pose-supervised 100-step models. Both selectors use the same 152 validation origins at stride 100, adding no training updates. Each family retains its own objective: NLL, response-mean MSE, and $L_{v_y}$ for our conditional-mean recursion; response MSE and $L_{v_y}$ for Chen24; state MSE for DeepDynamics. Validation retains physical-response history for the residual models and measured-state history for DeepDynamics; test initialization is common. These comparisons assess validation horizons within the implemented adaptations.

\subsection{Lateral Responses and Loss Ablations}
Table~\ref{tab:mean-accuracy} compares free simulation over the six test dynamic windows. With the common 100-step selector, increasing our training horizon from 1 to 100 reduces $v_y$ RMSE from 0.2496 to 0.0441 m/s and yaw-rate RMSE from 4.400 to 0.858 deg/s. Relative to physics, the hundred-step model reduces $v_y$, yaw-rate, and lateral-position RMSE by 85.7\%, 68.1\%, and 61.0\%, respectively. The improvement is largest between short and long rollouts, while the 50- and 100-step models have similar lateral-velocity accuracy.

Removing $L_{v_y}$ increases hundred-step $v_y$ RMSE from 0.0441 to 0.1347 m/s. With pose supervision, removal increases it from 0.0437 to 0.1176 m/s. These comparisons support direct supervision of the accumulated lateral-velocity error. Pose supervision reduces $Y$ RMSE from 0.543 to 0.518 m at $H_{\mathrm{tr}}=20$, from 0.543 to 0.501 m at 50, and from 0.483 to 0.452 m at 100. Its effect on $v_y$ is smaller and not uniformly favorable: at 50 steps, $v_y$ RMSE changes from 0.0448 to 0.0481 m/s. The pose-supervised hundred-step model gives the lowest mean errors in all four columns.

For the DeepDynamics adaptation, 100-step selection reduces dynamic-window $Y$ RMSE from 1.071 to 0.596 m and yaw-rate RMSE from 1.384 to 1.064 deg/s, with $v_y$ RMSE remaining near 0.047 m/s. Chen24 selection has a mixed effect: 100-step validation reduces yaw-rate and position errors, while one-step validation gives a lower $v_y$ RMSE, 0.2356 versus 0.2586 m/s. Thus a longer selector need not improve every response, and does not replace training on self-generated trajectories.

Figure~\ref{fig:scenario} illustrates the response and integrated motion during the lane change. The physical model overestimates the lateral-velocity excursions. The rollout model captures the direction changes and subsequent recovery, while the DeepDynamics adaptation also closely follows $v_y$. The heading and horizontal trajectory show the accumulated consequences of the response errors, complementing the instantaneous acceleration and yaw-acceleration curves.

\begin{table*}[!t]
\centering
\setlength{\tabcolsep}{3pt}
\caption{Full-record lane-change RMSE under two validation-selection horizons, using the same saved training trajectories. Learned entries show the mean $\pm$ sample standard deviation over three repetitions; all runs are initialized once and scored to the record end. This is one complete scene, not a six-record aggregate. Bold marks the lowest unrounded mean in each column.}
\label{tab:full-scene}
\begin{tabular}{lccccc}
\toprule
Model & $H_{\rm val}$ & $v_y$ (m/s) & $r$ (deg/s) & $Y$ (m) & $\phi$ (rad) \\
\midrule
Physics & -- & $0.2278$ & $1.353$ & $25.413$ & $0.1364$ \\
Chen24 & 1 & $0.1686\pm 0.0122$ & $2.122\pm 0.209$ & $14.676\pm 3.970$ & $0.1181\pm 0.0290$ \\
Chen24 & 100 & $0.1827\pm 0.0149$ & $2.039\pm 0.147$ & $6.926\pm 4.109$ & $0.0617\pm 0.0121$ \\
DeepDynamics lateral & 1 & $0.0302\pm 0.0046$ & $1.142\pm 0.262$ & $35.542\pm 15.931$ & $0.1489\pm 0.0523$ \\
DeepDynamics lateral & 100 & $0.0274\pm 0.0059$ & $0.869\pm 0.043$ & $21.728\pm 1.626$ & $0.0952\pm 0.0120$ \\
Rollout $H_{\rm tr}=1$ & 1 & $0.3067\pm 0.0242$ & $4.138\pm 0.345$ & $76.875\pm 20.971$ & $0.3933\pm 0.1044$ \\
Rollout $H_{\rm tr}=1$ & 100 & $0.1842\pm 0.0036$ & $2.218\pm 0.017$ & $6.123\pm 2.012$ & $0.0523\pm 0.0101$ \\
Rollout $H_{\rm tr}=100$ & 1 & $0.0258\pm 0.0023$ & $0.711\pm 0.002$ & $3.313\pm 0.490$ & $0.0136\pm 0.0027$ \\
Rollout $H_{\rm tr}=100$ & 100 & $0.0259\pm 0.0029$ & $\mathbf{0.709\pm 0.009}$ & $2.764\pm 1.804$ & $0.0130\pm 0.0064$ \\
Rollout $H_{\rm tr}=100$ + $L_{\rm pose}$ & 1 & $0.0255\pm 0.0014$ & $0.712\pm 0.003$ & $3.901\pm 1.373$ & $0.0162\pm 0.0067$ \\
Rollout $H_{\rm tr}=100$ + $L_{\rm pose}$ & 100 & $\mathbf{0.0246\pm 0.0009}$ & $0.713\pm 0.005$ & $\mathbf{2.146\pm 0.885}$ & $\mathbf{0.0097\pm 0.0028}$ \\
\bottomrule
\end{tabular}
\end{table*}

\subsection{Observation Updates and Continuous Simulation}
Accurate one-step prediction and accurate free simulation measure different behaviors. With measurement updates, our one-step-trained model gives $v_y$ RMSEs of 0.0117 and 0.0123 m/s for one- and hundred-step selection, respectively. In complete-scene free simulation, the corresponding errors are 0.3067 and 0.1842 m/s, and the $Y$ errors are 76.88 and 6.12 m (Table~\ref{tab:full-scene}). These paired results change only the retained checkpoint within each training run. A longer selector therefore avoids some poor recursive behavior, but hundred-step training further reduces full-scene $v_y$ RMSE to 0.0259 m/s and $Y$ RMSE to 2.76 m under the same hundred-step selection rule.

\begin{figure*}[!t]
\centering
\includegraphics[width=\textwidth]{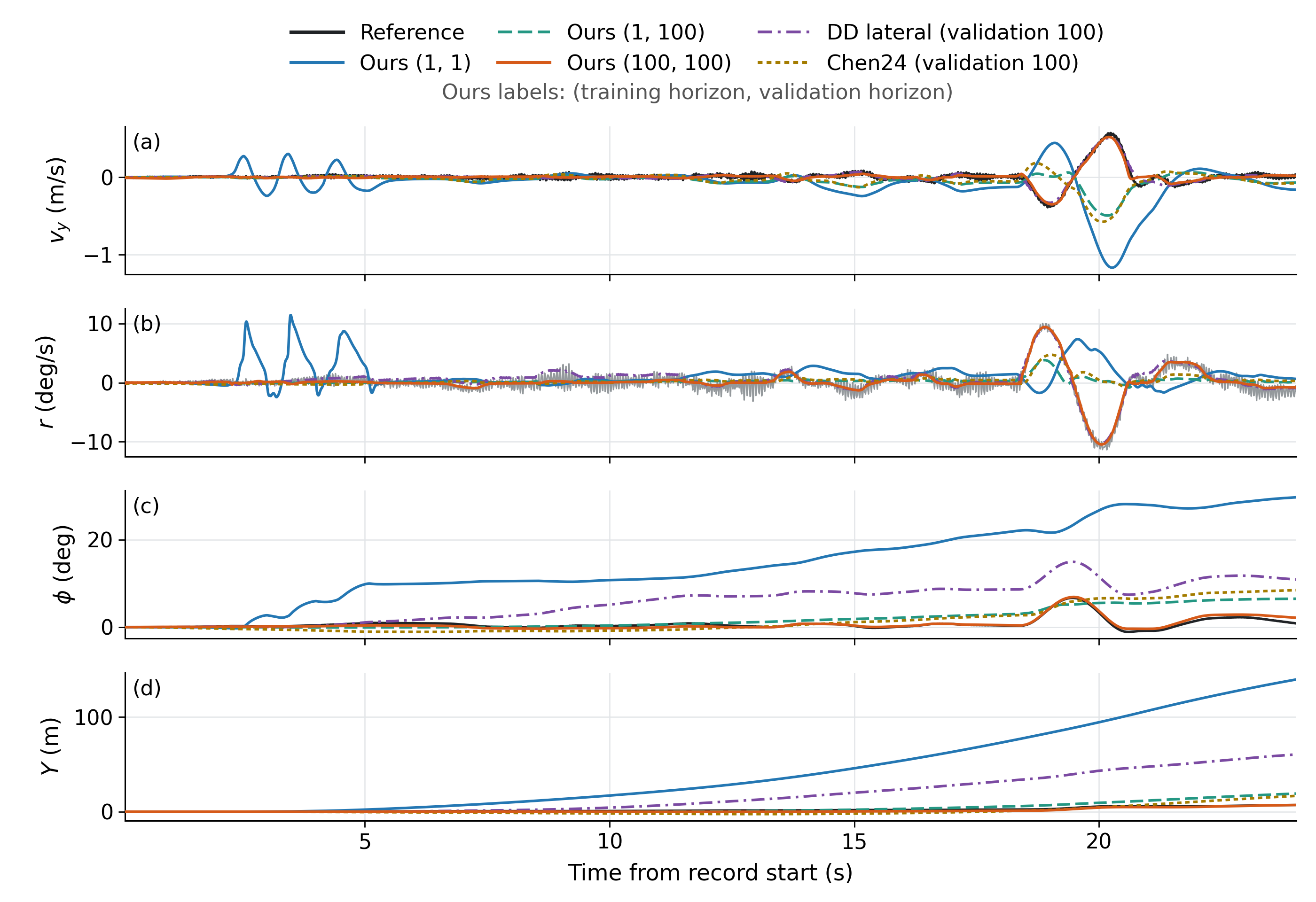}
\caption{Complete-scene free simulation with one measured initialization at 0.1 s and no later reset. Curves show a fixed training repetition; Table~\ref{tab:full-scene} summarizes all three. Training and validation horizons are labeled separately; Chen24 and DeepDynamics use 100-step validation. The entire recorded duration is shown, including the approach, lane change, and recovery.}
\label{fig:historical-drift}
\end{figure*}

Figure~\ref{fig:historical-drift} shows the artificial lateral response and accumulated pose error that can arise when a one-step-trained, one-step-selected model supplies its own future state. The hundred-step model more closely follows measured lateral velocity through the full record. Its pose-supervised variant reduces full-scene $Y$ RMSE to 2.15 m.

The DeepDynamics adaptation has lower observation-updated one-step $v_y$ error, 0.0046--0.0047 m/s, than our hundred-step model's 0.0118 m/s. Its complete-scene $Y$ error nevertheless reaches 35.54 m with one-step selection and 21.73 m with hundred-step selection. Its full-scene $v_y$ errors, 0.0302 and 0.0274 m/s, remain close to our rollout model's error; pose accumulation distinguishes the methods more strongly in this record. This comparison illustrates why small local state errors alone are insufficient for the integrated motion required by simulation. In predictive control, the latest state estimate reinitializes each prediction; free simulation must instead tolerate its own accumulated state and history errors.

\subsection{Propagated Response and Motion Uncertainty}
\begin{table}[!t]
\caption{Linearized mean $\pm3\sigma$ bands over one-second prediction windows for the predesignated hundred-step model used in Fig.~\ref{fig:scenario}. Coverage and full width are averaged within each record and then equally across six test records.}
\label{tab:coverage}
\centering
\begin{tabular}{lrr}
\toprule
Signal (unit) & Coverage (\%) & Mean width\\
\midrule
$v_y$ (m/s) & 94.0 & 0.221\\
$r$ (deg/s) & 98.2 & 5.031\\
$\phi$ (rad) & 68.5 & 0.0073\\
$Y$ (m) & 91.6 & 0.085\\
$a_y$ (m/s$^2$) & 99.6 & 7.831\\
$\dot r$ (deg/s$^2$) & 100.0 & 664.598\\
\bottomrule
\end{tabular}
\end{table}

The propagated bands in Fig.~\ref{fig:scenario} use the predesignated hundred-step model and the augmented-state linearization in~\eqref{eq:covariance}. Table~\ref{tab:coverage} evaluates this same model over 24 nonoverlapping one-second windows from the six test maneuvers, matching the training horizon. Each window is reinitialized from measured states and history and reference pose, with $P_0=0$, then predicted freely for 100 steps. Only complete windows are scored, excluding their initialization points. Empirical coverage is the fraction of reference samples within $\bar y_t\pm3\sigma_t$, and full width is $6\sigma_t$. Lateral-velocity and yaw-rate coverage is 94.0\% and 98.2\%; heading and lateral-position coverage is 68.5\% and 91.6\%. Acceleration and yaw-acceleration coverage is higher, with correspondingly broad intervals. These bands describe pointwise marginal spread.

\section{Discussion}
\textit{Lateral-velocity stability.} Persistent error in $a_y-u r$ accumulates in $v_y$. Rollout learning and direct state supervision reduce this effect in the evaluated records. Formal convergence requires analysis of the coupled state--history feedback, for example through contraction conditions~\cite{lohmiller1998contraction}. A longer validation horizon can improve checkpoint selection but cannot supply the gradients of multistep training.

\textit{Data and application scope.} Timing, coordinate conventions, bias, and filtering can make recorded velocities inconsistent with integrated responses. Independent pose measurements would strengthen the reconstructed $Y/\phi$ evaluation. External trials are still needed. Real-time execution and closed-loop platform tracking remain to be assessed.

\textit{Uncertainty and temporal dependence.} Empirical band coverage is 68.5\% for heading and 98.2\% for yaw rate. The lower heading coverage suggests that the propagated heading uncertainty may be underestimated. Persistent yaw-rate bias or temporal dependence in the residuals can produce heading errors that exceed the propagated band. Linearization may also contribute. The learned covariance includes unresolved response variability, measurement noise, and model mismatch. For $\dot r=(r_{t+1}-r_t)/\Delta t$, adjacent-rate covariance and the $\Delta t^{-2}$ factor can produce wide acceleration bands, so coverage must be read with width~\cite{gneiting2007calibration}. A 100 Hz update rate alone does not determine disturbance correlation. Residual temporal dependence should therefore be assessed before using the covariance to inject physical disturbances, potentially through an identified colored-noise model~\cite{sarkka2019sde}.

\section{Conclusion}

This paper presented a physics-guided probabilistic lateral motion prediction model for vehicle-in-the-loop simulation. By combining a two-degree-of-freedom vehicle model with recurrent probabilistic residual learning, the proposed model predicts correlated corrections to lateral acceleration and yaw rate and integrates the corrected responses to generate horizontal lateral position and heading. This formulation connects lateral-response prediction with the integrated motion required by the target simulation platform.

To improve response fidelity during recursive motion prediction, multistep rollout training was introduced to expose the model to its own evolving states and response histories. Direct supervision of accumulated lateral-velocity error helps constrain the effects of persistent response imbalance and reduces prediction errors during free-running simulation. Experimental results demonstrate improvements in both lateral-response and integrated-motion prediction. In particular, the proposed hundred-step rollout model achieves mean lateral-velocity and lateral-position RMSEs of 0.0441 m/s and 0.483 m, respectively, over the evaluated test maneuvers. Compared with the physical model, these correspond to reductions of 85.7\% and 61.0\%, respectively. The proposed model also reduces lateral-position RMSE by 19.0\% compared with the implemented Deep Dynamics lateral adaptation.

Furthermore, the conditional response covariance is propagated through an augmented-state linearization to characterize uncertainty in lateral responses and integrated vehicle motion. The resulting framework provides both predicted motion trajectories and associated uncertainty intervals. Future work will focus on improving uncertainty calibration, investigating temporal dependence in response disturbances, and evaluating real-time execution and closed-loop tracking performance on the vehicle-in-the-loop platform.

\FloatBarrier

\end{document}